\pdfoutput=1

\documentclass[11pt]{article}

\usepackage{acl}

\usepackage{times}
\usepackage{latexsym}

\usepackage[T1]{fontenc}

\usepackage[utf8]{inputenc}

\usepackage{microtype}
\usepackage{graphicx}
\usepackage{amsmath}

\usepackage{amssymb}
\usepackage[ruled,linesnumbered]{algorithm2e}
\usepackage{multirow}
\usepackage{color}
\usepackage{booktabs}
\usepackage{array}
\usepackage{makecell}
\usepackage{caption}
\usepackage{subfigure}
\usepackage{marvosym}
\title{Few-Shot Nested Named Entity Recognition}

\author{Hong Ming$^1$ \and Jiaoyun Yang$^1$\textsuperscript{\Letter}  \and Lili Jiang$^2$ \and Yan Pan$^1$ \and Ning An$^1$ \\
        $^1$Hefei University of Technology  \\ 
        $^2$Umeå University   \\ 
        \texttt{\{ming\_hong, yanpan\}@mail.hfut.edu.cn}  \\ \texttt{jiaoyun@hfut.edu.cn} \\ \texttt{lili.jiang@cs.umu.se} \\ \texttt{ning.g.an@acm.org}
        }

\begin{document}
\maketitle
\begin{abstract}
While Named Entity Recognition (NER) is a widely studied task, making inferences of entities with only a few labeled data has been challenging, especially for entities with nested structures. Unlike flat entities, entities and their nested entities are more likely to have similar semantic feature representations, drastically increasing difficulties in classifying different entity categories in the few-shot setting. 
Although prior work has briefly discussed nested structures in the context of few-shot learning, to our best knowledge, this paper is the first one specifically dedicated to studying the few-shot nested NER task.
Leveraging contextual dependency to distinguish nested entities, we propose a Biaffine-based Contrastive Learning (BCL) framework. We first design a Biaffine span representation module for learning the contextual span dependency representation for each entity span rather than only learning its semantic representation. We then merge these two representations by the residual connection to distinguish nested entities. Finally, we build a contrastive learning framework to adjust the representation distribution for larger margin boundaries and more generalized domain transfer learning ability. We conducted experimental studies on three English, German, and Russian nested NER datasets. The results show that the BCL outperformed three baseline models on the 1-shot and 5-shot tasks in terms of $F_1$ score.

\end{abstract}

\section{Introduction}
\begin{figure}[h]
  \centering  
  \subfigure[Example of a sentence with nested entities.]{
    
    \begin{minipage}[t]{0.98\linewidth}
      \centering 
    \includegraphics[width=1\linewidth]{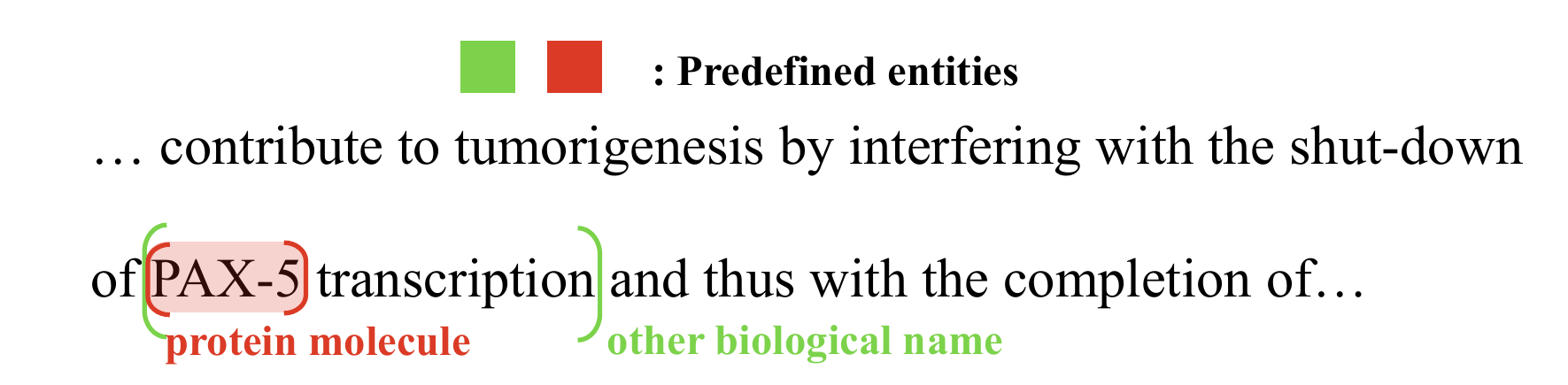} 
  
    \end{minipage}
  }

  \subfigure[Span representations from BERT model (left) and BCL model (right). ]{
    \begin{minipage}[t]{0.98\linewidth}
      \centering 
    \includegraphics[width=1\linewidth]{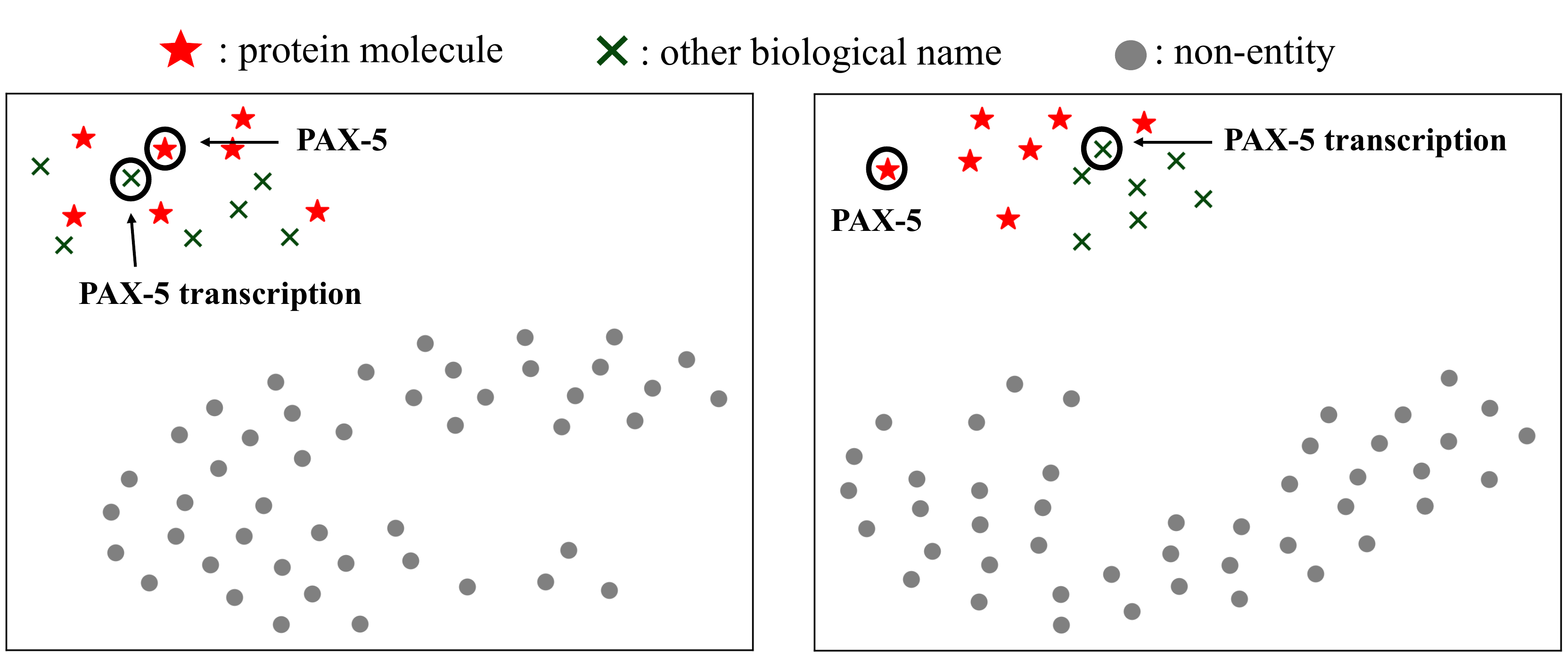} 
  
    \end{minipage}

   }
  \caption{Illustration of nested entities (a) and the distribution of entity span representations obtained from BERT (b left) and our BCL  model (b right). The samples are extracted from the GENIA dataset. The red star, green cross, and grey circle symbols denote ``protein molecule'' category entities, ``other biological name'' category entities, and non-entities, respectively. In each sample, a ``protein molecule'' category entity is nested in an ``other biological name'' category entity.  Representations obtained from our BCL model exhibit more distinct clusters.}
  \label{figure2_hard_transfer}   
  \end{figure}

As a fundamental task in the Natural Language Processing (NLP), Named Entity Recognition (NER) is mainly about extracting the boundaries and categories of entities in the given sentences, which can provide helpful information for down-stream tasks like text classification \citep{lee2018abusive}, event detection \citep{popescu2011extracting}, question answering \citep{lee2007fine}, and more. NER tasks are normally divided into the flat NER task and the nested NER task \citep{finkel2009nested}. The nested NER task is more challenging since an entity could be part of another, resulting in a nested structure. Figure \ref{figure2_hard_transfer} (a) illustrates an example of nested entities from the GENIA dataset \citep{kim2003genia}, where entity ``PAX-5'' is nested in the entity ``PAX-5 transcription''. 
While the former belongs to the ``protein molecule'' category, the latter belongs to the ``other biological name'' category.
Nested entities commonly exit. About 53.9\% entities are nested in the GENIA dataset.

Researchers have made significant progress on the nested NER task by applying deep learning models \citep{tan2021sequence, yuan2021fusing, shen2021locate, ma2022decomposed, wang2021enhanced, fu2021nested}, including various pre-trained models, particular sequence labeling strategies, or graph convolutional networks. These well-performed deep learning models require extensive and high-quality manually labeled data. However, these data are often not readily available due to the lack of domain knowledge from human annotators, the high cost of annotating large-scale data, or the restriction of privacy and security \citep{lu2020learning}. This motivates us to explore the few-shot nested NER task that requires recognizing unlabeled instances (aka., query set) according to only a few labeled samples (aka., support set).

For the few-shot flat NER task, considering the limited information gained from very few labeled samples, researchers have developed various types of methods, including meta-learning methods, prompt tuning methods, and contrastive learning methods \citep{hofer2018few, ma2021template,das2021container,wang2021enhanced,ma2022decomposed}, to learn information from other rich resource domains. As the entities are not overlapped in the few-shot flat NER task, these methods usually apply the sequence labeling strategy to mark each entity's boundaries and category. When it comes to the few-shot nested NER task, the issue of nested structure among entities has to be addressed.

\begin{figure}[h]
  \centering
  \includegraphics[width=0.48\textwidth]{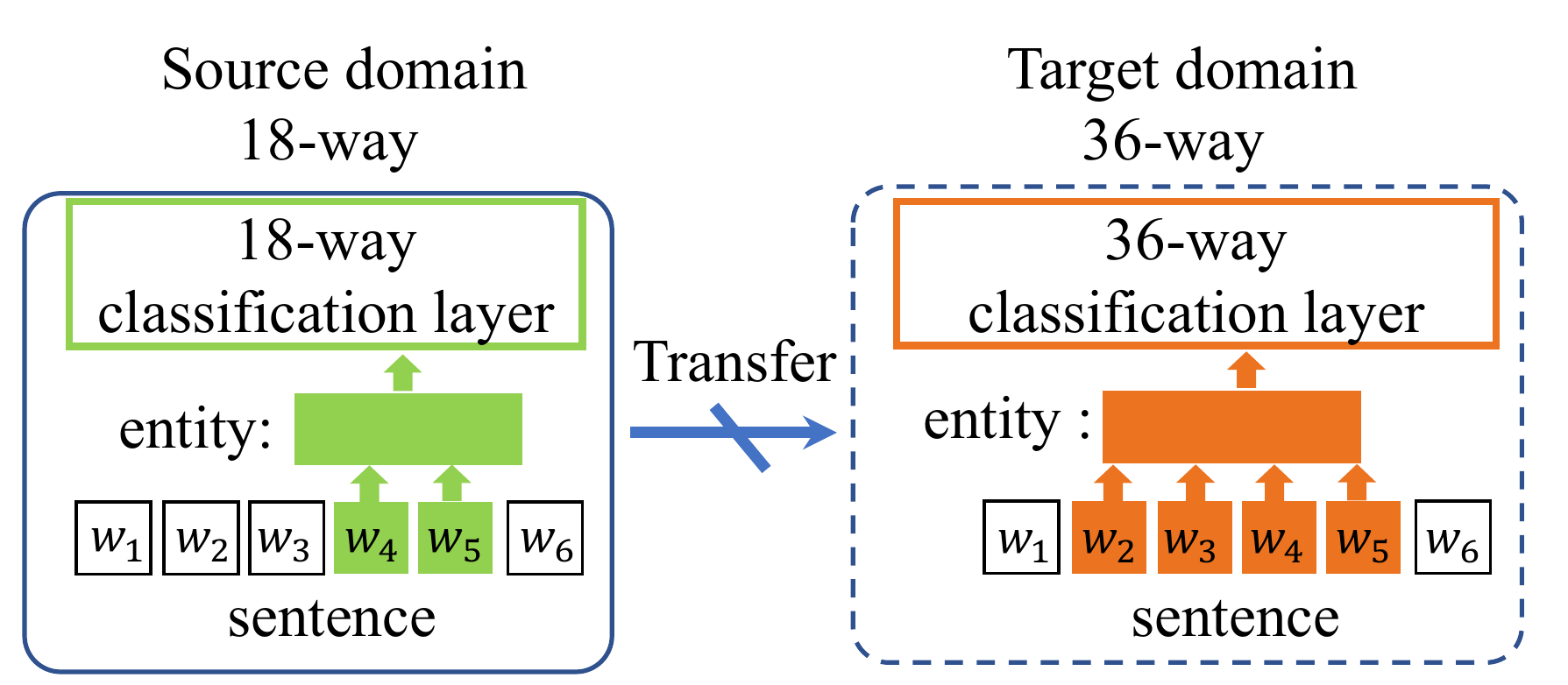} 
  \caption{Illustration of label discrepancy when applying transfer learning.}
  \label{figure3_label_discrepancy}
\end{figure}

This paper simultaneously addresses the double challenges of nested entity structure and very few training samples in the few-shot nested NER task.
For the challenge of nested entity structure, we apply entity span representations to cover all the word combinations. However, it raises an issue that the entity spans and their nested entity spans are more likely to obtain similar semantic feature representations due to the shared words. We generate Figure \ref{figure2_hard_transfer} (b left) to illustrate the distribution of the semantic representations of spans and their nested spans by performing the dimension reduction method t-SNE \citep{van2008visualizing} on the span representations obtained from the BERT \citep{devlin2018bert} model. We can find that entity spans and their nested entity spans are mixed together, resulting in hard to distinguish. For the challenge of very few training samples, we apply the transfer learning framework to learn information from other rich resource domains. However, there is a label-discrepancy issue when classifying entity span representations into entity categories. As shown in Figure \ref{figure3_label_discrepancy}, though learning information from other rich resource domains can partially overcome the shortage of training samples in the target domain, the number of label categories in the source domain and the target domain is not consistent, i.e., the label-discrepancy issue. This makes classifiers trained on the resource domain hard to transfer to the target domain.

To overcome the aforementioned in the few-shot nested NER task, we propose a Biaffine-based Contrastive Learning (BCL) framework. It is observed that entity spans and their nested entity spans could have different dependency patterns and the dependency between words within an entity span and words outside this entity span can help recognize entities. Motivated by this, we propose to learn contextual span dependency representation using a Biaffine span representation module instead of only learning the semantic representation for each entity span.

To illustrate the effect of the dependency representation, we merge the dependency representation into the semantic representation on the entity spans in Figure \ref{figure2_hard_transfer} (b left) and apply t-SNE to obtain the span representation distribution, shown in Figure \ref{figure2_hard_transfer} (b right). Compared with Figure \ref{figure2_hard_transfer} (b left), spans in Figure \ref{figure2_hard_transfer} (b right) exhibit more distinct clusters.  
After obtaining the span representation, we further apply the circle loss \citep{sun2020circle} in the contrastive learning framework to adjust the representation distribution by enhancing the similarity of spans of the same entity category and reducing the similarity of spans of different entity categories. In this way, the final entity labels can be achieved by similarity measurement. Therefore, the number of label categories in the source and target domains is not required to be consistent, which enhances the capability of transfer learning on the few-shot problem.

Our main contributions are as follows:
\begin{itemize}
  \item We propose a Biaffine-based contrastive learning framework to tackle the few-shot nested NER task. Although prior work has briefly discussed nested structures in the context of few-shot learning, to our best knowledge, this paper is the first one specifically dedicated to studying the few-shot nested NER task.
  \item We design a Biaffine span representation module for learning contextual span dependency representation and further merge the dependency representation with the semantic representation by the residual connection for distinguishing nested entities.
  \item We conduct experiments on three nested NER datasets from three different languages. The results show improvements in our proposed  BCL framework over the existing few-shot learning methods in terms of $F_1$ score.
\end{itemize}

The rest of this paper is organized as follows. Section 2 states the few-shot nested NER task and existing few-shot learning methods in NER. Section 3 details our BCL model for the few-shot nested NER task. Following the experimental study in Section 4, we conclude this paper in Section 5.

\section{Preliminaries}
This section first introduces the few-shot nested NER task and then discusses the current few-shot learning methods in NER.

\subsection{Problem Statement}
We formulate the few-shot nested NER task as an entity span classification problem to cover all possible word combinations. Therefore, given an input sentence $x\in\mathcal{X}$ with $n$ tokens, denoted by $x=\{w_1,\ldots,w_n\}$, we generate an entity span set, where each span is a span of tokens $\{w_p,\ldots,w_q\} \ (1  \leqslant  p  \leqslant  q  \leqslant  n)$\normalsize. Then, we learn a classification model to map each span into a category label in the label set $E_\mathcal{X}$. If we set the task as a $K$-shot task, then the number of span labels in $\mathcal{X}$ is limited to $K$. As we also apply the transfer learning framework, the formal descriptions are as follows.

Suppose $\mathcal{X}_i$ and $\mathcal{X}_j$ are the sentence sets in the source domain $i$ and the target domain $j$, respectively. $\mathcal{X}_j$ is further divided into a support set $\mathcal{X}_j^{spt}$ and a query set $\mathcal{X}_j^{qry}$. The few-shot nested NER task first trains a model on $\mathcal{D}_i=\left\{\mathcal{X}_i, \mathcal{Y}_i\right\}$, where  $\mathcal{Y}_i$ is the corresponding span labels of $\mathcal{X}_i$. Then it makes adaptations on $\mathcal{X}_j$, i.e., it first fine-tunes the model on $\mathcal{D}_j^{spt}=\{\mathcal{X}_j^{spt}, \mathcal{Y}_j^{spt}\}$ and then predicts the span labels for $\mathcal{D}_j^{qry}=\{\mathcal{X}_j^{qry}\} $, where $\mathcal{Y}_j^{spt} $ is the corresponding span labels of $\mathcal{X}_j^{spt} $ and each entity category in $\mathcal{X}_j^{spt} $ only contains $K$ entities. There are two fundamental rules for datasets in the domain $i$ and $j$:

\begin{itemize}
    \item The entity categories in $\mathcal{D}_i$ are different from categories in $\mathcal{D}_j: E_{\mathcal{X}_i}\ \cap\ E_{\mathcal{X}_j}=\ \varnothing$.
    \item The entity categories in $\mathcal{D}^{spt}$ are the same as categories in $\mathcal{D}^{qry}$, but sentences appearing  in $\mathcal{D}^{spt}$ will NEVER appear in $\mathcal{D}^{qry} : E_{\mathcal{X}^{spt}}=E_{\mathcal{X}^{qry}},\ \mathcal{X}^{spt}\cap\ \mathcal{X}^{qry}=\ \varnothing$.
\end{itemize}

\begin{figure*}[h]
  \centering
  \includegraphics[width=0.98\textwidth]{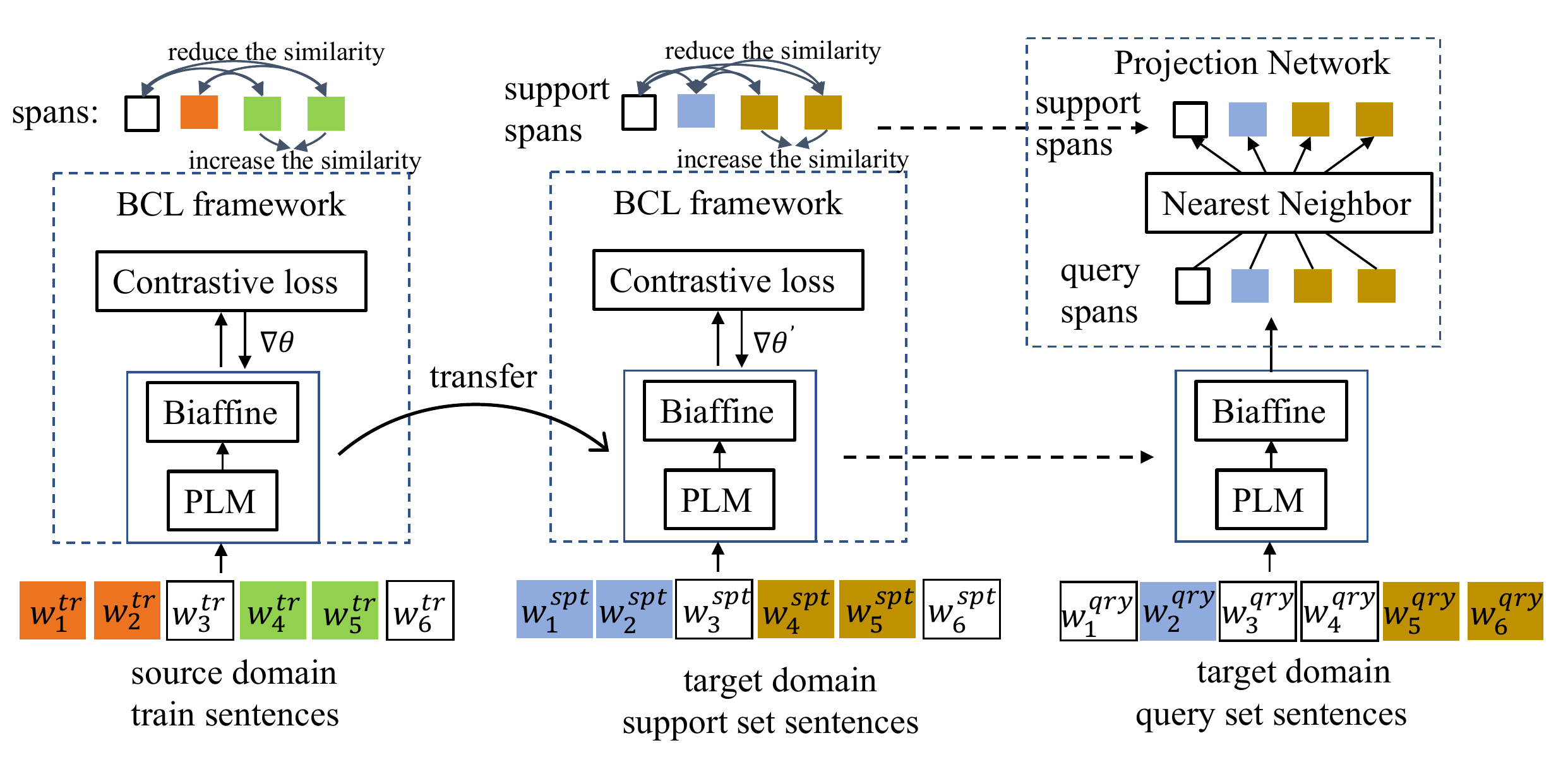} 
  \caption{The overall framework of our proposed BCL and the learning procedures. In BCL, words are first encoded as semantic representations by the Pre-trained Language Model (PLM) and then converted into span representations in the Biaffine module. After that, the representations are adjusted by the contrastive loss to reduce the similarity of the spans in different categories and increase the similarity of spans in the same category. When learning the model, we first train BCL($\theta $) in the source domain by using the circle loss as the contrastive loss. After training, we transfer our BCL to the target domain. We utilize the support set with $K$-shot samples to fine-tune the BCL($\theta ' $) and then use the model to make inferences in the query set with the nearest neighbor algorithm.}
  \label{model_framework}
\end{figure*}

\subsection{Few-shot NER Frameworks}
As there are few previous works dedicated to the few-shot nested NER task, we review some relevant frameworks applied in the few-shot flat NER task. 

Most of the few-shot NER methods are based on the transfer learning framework, which aims to train models from the source domain and adapt them to the target domain.
The target domain $j$ only contains a few labeled sentences, while the source domain $i$ can have plenty of labeled data \citep{yang2021few, ding2021few}. Traditional transfer learning methods usually use the whole data in the source domain to train a sophisticated model and then fine-tune it in the target domain. This is more likely to cause overfitting in the few-shot setting. Therefore, various strategies are designed for better learning in the source domain.

Meta-learning-based methods sample many few-shot subtasks in the source domain to make the model fit the few-shot setting, which shows to be effective in the few-shot tasks. This type of method could be further categorized as metric-based methods that need to calculate the distance between training samples and test samples \citep{snell2017prototypical}, memory-based methods that need to store and maintain input representations in an external memory \citep{florez2019learning}, and optimization-based methods that need to use additional steps to optimize the meta tasks \citep{finn2017model,li2020few}.

Considering the difficulty of locating useful information in the few-shot setting, prompt-based methods have been put forward to strengthen the few-shot learning strategies. Prompt-based methods design the template of additional information to  help model learning from a few resources \citep{liu2021pre, hou2022inverse,hu2021knowledgeable,ma2021template}. 

When applying these transfer learning methods, an issue needed to consider is that the number of categories in the source and target domains may be inconsistent. This leads to a label dependency problem that requires retraining the model when transferring from the source domain to the target domain \citep{hou2019few}.

Recently, contrastive learning methods are proposed to focus on the similarity of samples in the same or different categories \citep{gao2021simcse, das2021container}. Owing to no need to set the classification layer and retrain on the new tasks, the contrastive learning methods could easily adapt to other few-shot NER tasks. This is also why we apply the contrastive learning framework to tackle the few-shot nested NER task.

\section{Methodology}
In this section, we first explain the framework of the proposed Biaffine-based Contrastive Learning (BCL) model. Afterward, we detail the Biaffine span representation module for generating span representations.

\subsection{Biaffine-based Contrastive Learning Framework}

Figure \ref{model_framework} illustrates the overall framework of BCL and the learning procedures. 

BCL consists of three modules: the pre-trained language model, Biaffine span representation, and contrastive optimization. We first utilize $\rm {BERT}_{base\_multilingual}$  \citep{devlin2018bert} as the pre-trained language model to generate word semantic embedding vectors. Then, the embedding vectors are further converted to the span representations in the Biaffine span representation module, which will be detailed in the next section.
and then apply a BiLSTM layer to further learn contextual features.

In the contrastive optimization module, we use the circle loss to adjust the representation distribution by increasing the similarity between positive samples and reducing the similarity between negative samples. The loss function is as follows:
\begin{equation}
  \mathcal{L}_s = log(1 +  sim(s,s^+)  * sim(s,s^-) )
\end{equation}
Where  $s$ denotes a specific span, $s^+$ denotes all positive samples with the same category as $s$, and $s^-$ denotes all negative samples with different categories from $s$. $sim(s,s^+)$ denotes the total similarity between $s^+$ and $s$, and $sim(s,s^-) $ denotes the total similarity between  $s^-$ and $s$, which can be calculated by:
\begin{equation}
  \label{equ_simpos}
  sim(s,s^+) = \sum_{s_i^+}^{s^+}   e^{-\tau \ * \ similarity(s, s_i^+) }
\end{equation}
\begin{equation}
  sim(s,s^-) = \sum_{s_i^-}^{s^-}  e^{\tau \ * \ similarity(s, s_i^-)}
\end{equation}
Where $similarity(a,b)$ denotes the cosine similarity of $a$ and $b$, $\tau$ is the temperature \citep{wang2021understanding}. 

\subsection{Biaffine Span Representation}

\begin{figure}[h]
    \centering
    \includegraphics[width=0.48\textwidth]{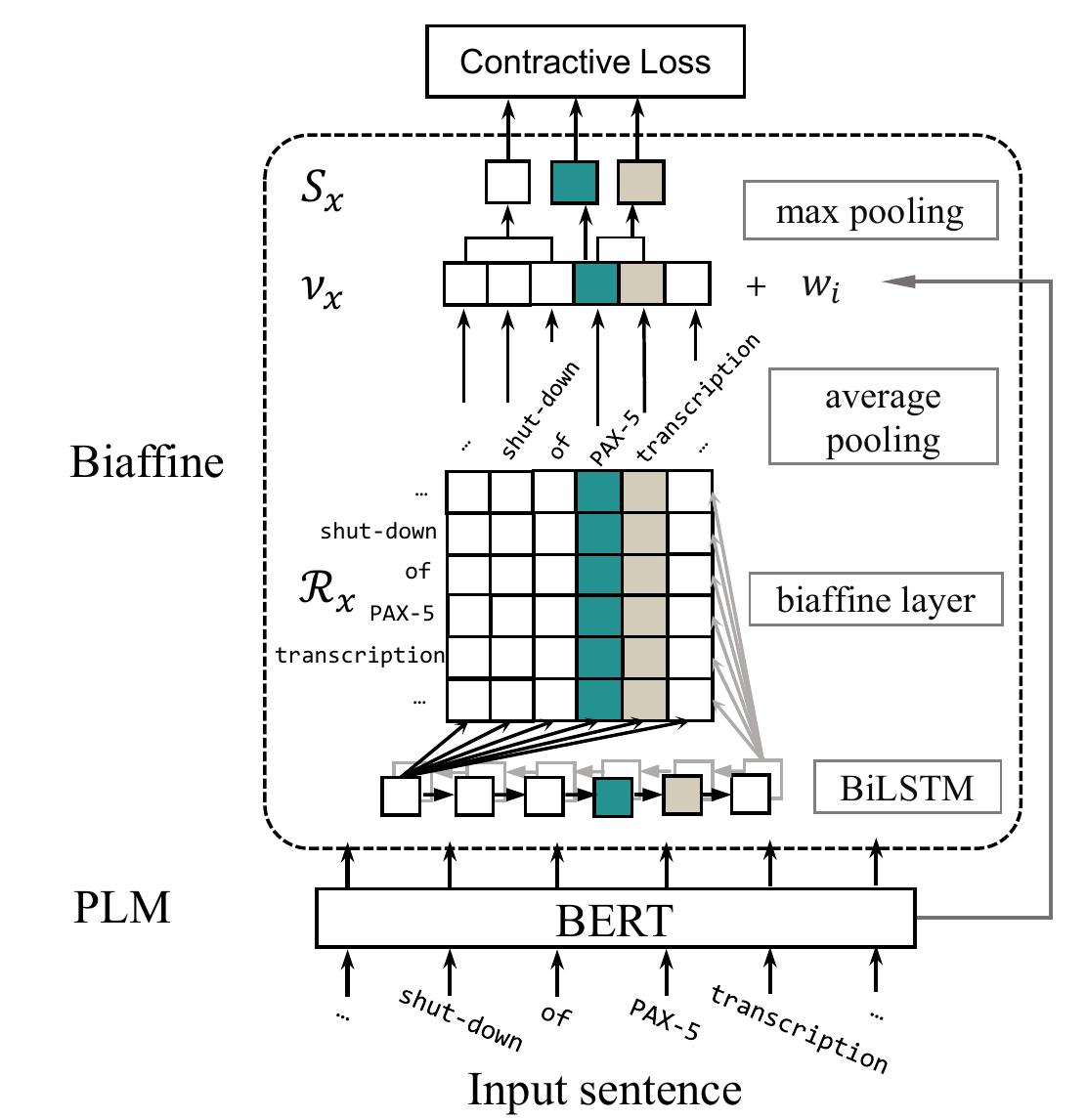} 
    \caption{The architecture of the Biaffine span representation module. After embedding words with BERT, the module first utilizes a BiLSTM layer to learn contextual features. Then, it applies a Biaffine layer and an average pooling layer to learn the contextual dependency representation for each word. After that, the dependency representation is merged with the semantic representation from BERT by the residual connection. Finally, the module utilizes a maximum pooling layer to obtain the span representation.}
\label{figure3}
\end{figure}

Figure \ref{figure3} illustrates the architecture of the Biaffine span representation module. The module receives word semantic embedding vectors from the PLM module and then utilizes a BiLSTM layer to learn contextual features. After that, it applies a Biaffine layer \citep{yu2020named, dozat2016deep} and an average pooling layer to learn the contextual dependency representation for each word.

The calculation in the Biaffine layer could be represented by:
\begin{equation}
    \mathcal{R}_x = \ {\mathcal{H}_f}^\intercal {\ \mathbf{U}}_x^1\ \mathcal{H}_b \ + \ {(\mathcal{H}_f\ \oplus\ \mathcal{H}_b)}^\intercal \ {\mathbf{U}}_x^2 \ + \ \mathbf{b} 
\end{equation}
Where $r$ is a low-dimensional space, $\mathcal{H}_f \in \mathbb{R}^n$ is the forward word representation obtained from BiLSTM, $\mathcal{H}_b \in \mathbb{R}^n$ is the backward word representation obtained from BiLSTM, $\mathbf{U}_x^1$ is a $h\times r\times h$ tensor with $h$ as the dimension of $\mathcal{H}_f$ or $\mathcal{H}_b$, $\mathbf{U}_x^2$ is a $2h\times r$ matrix, and $b$ is the bias. 

In Figure \ref{figure3}, $\mathcal{R}_x$ is represented as a $n \times n$ matrix with $n$ as the length of sentence $x$. Each element in the matrix is a dependency feature vector in the low-dimensional space, i.e., $\mathcal{R}_x(i,\ j)$ means the Biaffine dependency score of the $i^{th}$ word on the $j^{th}$ word. 

The average pooling layer is used to calculate the average Biaffine dependency score for each word with the other words in the sentence. Then the module merges this dependency presentation with the semantic representation from BERT by the residual connection to obtain the final word representation. The calculation is as follows:
\begin{equation}
  \upsilon_i=\ \frac{1}{n}  \sum_{j = 1}^{n}  \mathcal{R}_x(i,\ j) + w_i
\end{equation}
Where $\upsilon_i$ is the final word representation, $w_i$ is the word semantic representation obtained from BERT.

Finally, for a span ${s}_x$ containing words $\{w_p,\ldots,w_q\} \ (1  \leqslant  p  \leqslant  q  \leqslant  n)$, we adopt the max pooling  to extract the max vector within each span to highlight the important features of each word in this span.
\begin{equation}
{s}_x=\ Max( \upsilon_p, \ldots, \upsilon_q ) 
\end{equation}

\section{Experiments}
This section presents our experimental study to evaluate our proposed BCL framework on the few-shot nested NER task. We first introduce the datasets and baselines used in our experiments. We then describe the experiment procedures and settings. After that, we present and discuss the experimental results.

\subsection{Datasets and Baselines}

\begin{table}[]
  \scriptsize
  \caption{Datasets used in experiments}
\label{tab1}
\centering
\begin{tabular}{ccccc}
  \toprule[1pt]
  Dataset    & language  & Types & Sentences & Entities/Nest entities \\ 
  \midrule[1pt]
  GENIA      & English & 36    & 18.5k  & 55.7k / 30.0k\\
  GermEval  & German & 12    & 18.4k     & 41.1k / 6.1k  \\
  NEREL   &  Russian & 29    & 8.9k   & 56.1k / 18.7k    \\ 
  FewNERD & English & 66    & 188.2   & 491.7k / -   \\ 
  \bottomrule[1pt]    
\end{tabular}
\end{table}

As shown in Table \ref{tab1}, we use three nested NER datasets and a flat NER dataset in our experiments including GENIA \footnote{\url{http://www.geniaproject.org/genia-corpus}} \citep{kim2003genia}, GermEval\footnote{\url{https://sites.google.com/site/germeval2014ner/data}} \citep{benikova2014nosta},NEREL \footnote{\url{https://github.com/nerel-ds/NEREL}} \citep{loukachevitch2021nerel} and FewNERD \footnote{\url{https://ningding97.github.io/fewnerd/}} \citep{ding2021few}. The former three datasets are nested NER datasets, and the last one is a flat NER dataset. All these datasets are publicly available under the licenses of CC-BY 3.0, CC-BY 4.0, CC-BY 2.5, and CC-BY-SA 4.0, respectively. We have manually checked to guarantee these datasets are without offensive content and identifiers. Following the data statement, we use the FewNERD dataset as the source domain dataset and others as the target domain dataset. 

We compare the proposed BCL framework with the following baseline models:

\begin{itemize}
  \item CONTaiNER \citep{das2021container} is a contrastive learning method focusing on word embeddings. It assumes that the embeddings follow the Gaussian distributions and use KL-divergence to measure the similarity between words. The loss function is the average of similarities between positive samples dividing similarities between all samples. We adapt this method to handle the nested NER task.
\item ProtoNet \citep{snell2017prototypical} is a metric-based meta-learning method for the few-shot NER task. It focuses on computing distances between the center points of training samples (prototypes) and test samples. The model uses the BERT to get word embeddings. It further applies the contrastive learning framework so the model could be trained in the source domain and adapted to a totally different target domain. The loss function is the same as CONTaiNER.
\item NNShot \citep{yang2020simple} is a meta-learning model that determines the current token category according to token-leave distance from labeled training sets. And similar to ProtoNet, the model also uses the BERT to get word embeddings and applies the contrastive learning framework. Also, the loss function is the same as CONTaiNER. Besides, since we utilize the entity spans to represent the entity rather than sequence labeling, we do not need the CRF (Conditional Random Field) layer to label the tokens. So the StructShot method mentioned in yang \citep{yang2020simple} will not be used in our experiment.
	
\end{itemize}

\begin{table*}[]
  \small
  \centering
  \caption{ Performance on GENIA, GermEval, and NEREL nested NER datasets with 1-shot and 5-shot settings (\%).}
  \label{tab2}
  \begin{tabular}{c|cc|cc|cc}
    \toprule[1pt]
  \multirow{2}{*}{Model} & \multicolumn{2}{c|}{GENIA}           & \multicolumn{2}{c|}{GermEval}        & \multicolumn{2}{c}{NEREL}           \\ 
                         & \multicolumn{1}{c|}{1-shot} & 5-shot & \multicolumn{1}{c|}{1-shot} & 5-shot & \multicolumn{1}{c|}{1-shot} & 5-shot \\ \midrule[1pt]
    NNShot               & \multicolumn{1}{c|}{23.51 $\pm $ 1.72}  & 27.91 $\pm $ 1.46      & \multicolumn{1}{c|}{28.58 $\pm $ 6.76}      & 41.26 $\pm $ 2.50      & \multicolumn{1}{c|}{38.58 $\pm $ 1.30}      & 46.54 $\pm $ 1.93     \\ 
    ProtoNet             & \multicolumn{1}{c|}{12.46 $\pm $ 1.19}      & 15.55 $\pm $ 0.60      & \multicolumn{1}{c|}{19.05 $\pm $ 1.71}      & 28.59 $\pm $ 2.32      & \multicolumn{1}{c|}{17.76 $\pm $ 1.78}      & 23.16 $\pm $ 3.19     \\ 
    CONTaiNER            & \multicolumn{1}{c|}{27.63 $\pm $ 1.27}      & 44.77 $\pm $ 1.06     & \multicolumn{1}{c|}{33.18 $\pm $ 6.03}      & 42.38 $\pm $ 2.61      & \multicolumn{1}{c|}{35.23 $\pm $ 2.31}      & 53.55 $\pm $1.14      \\ 
  BCL                   & \multicolumn{1}{c|}{33.71 $\pm $ 1.75}  & 46.06 $\pm $ 1.22 & \multicolumn{1}{c|}{39.56 $\pm $ 5.69}  & 47.07 $\pm $ 2.94& \multicolumn{1}{c|}{44.47 $\pm $1.60}  & 58.95 $\pm $ 1.64\\ 
  \bottomrule[1pt]
  \end{tabular}
  \end{table*}

\subsection{Training and Testing}
In the training procedure, we utilized the FewNERD dataset, which could be decomposed into the inter- and intra-domain parts \citep{ding2021few}. 
We randomly sample 5-way 5-shot subtasks from the FewNERD inter-domain subset for training, among which 10,000 subtasks as the training set and 500 subtasks as the validation set. We use the validation set to validate the framework every 1000 subtasks during the training procedure.

In the testing procedure, we first sampled several sentences in the target domain dataset as the fine-tune dataset. When sampling, we limited the number of entities in each entity category to $k$. Some sentences contain more than one entity. Thus, some entity categories may have more than $k$ entities after the sampling procedure. After that, we made the test dataset using the left sentences which were not in the fine-tune dataset. We then fine-tuned our model on the fine-tune dataset and tested our model on the test dataset.  Here we chose 1-shot and 5-shot as the settings of the few-shot fine-tune dataset. Note that the model was trained in the FewNERD dataset and would not be retrained in the target dataset even if the target dataset was from other languages. For the GENIA dataset, we dropped entity categories with a number of entities less than 50. For the NEREL and GermEval datasets, the sampled datasets are from the given test part of the original datasets.

\subsection{Experiment Settings}
\label{section:Implementation details}
To encode words in different languages into vectors, we use the Pre-trained Language Model (PLM) $\rm {BERT}_{base\_multilingual}$ which has 12 heads of attention layers and 768 word-embedding dimensions. The BiLSTM layer has one hidden layer whose hidden size is 512. The initial states $h_0$ and $c_0$ are initialized randomly. The dimension in the Biaffine layer is 256. We normalize the results after the Biaffine layer and utilize a dropout of 0.2. The temperature  $\tau$ is set to 10, and the bias $\lambda$ is set to 30.  We implemented BCL with Huggingface Transformer 4.21.1 and PyTorch 1.12.1. The experiments were performed on Nvidia Tesla V100 GPUs.

\subsection{Experimental Results}

Table \ref{tab2} shows the average $F_1$ results over ten iterations with different random seeds on GENIA, GermEval, and NEREL nested NER datasets. 
Compared to NNShot, ProtoNet, and CONTaiNER, BCL obtains the best performance, achieving an average increase of 9.02\%, 22.82\%, 7.23\% on the 1-shot task and 12.12\%, 28.14\%, 3.79\% on the 5-shot task in terms of $F_1$ score, respectively. 

Among the four models, ProtoNet performs the worst, the reason may be that ProtoNet applies a metric-based learning strategy and finds center points as prototypes. In the few-shot setting, there is a distribution shift resulting in the center point of very few samples could not stand for the whole entity category. Among the three datasets, these four models perform worse on GENIA than on GermEval and NEREL. This is because GENIA contains the most nested entities, which increases the difficulty of recognizing entities.

  \subsection{Experimental Analysis}
  \subsubsection{Ablation Study}
  \begin{table}[]
    \small
    \centering
    
    \caption{ Ablation study on the GENIA dataset (\%).}
    \label{tab3}
    \begin{tabular}{l|cc}
      \toprule[1pt]
    \multirow{2}{*}{}       & \multicolumn{2}{c}{GENIA}           \\ 
                            & \multicolumn{1}{c|}{1-shot} & 5-shot \\ \midrule[1pt]
BCL & \multicolumn{1}{c|}{33.71 $\pm $ 1.75}  & 46.06 $\pm $ 1.22\\
 \quad w/o residual connection & \multicolumn{1}{c|}{31.95 $\pm $ 1.27}       &44.77 $\pm $   1.06     \\
  \quad w/o Biaffine layer      & \multicolumn{1}{c|}{31.98 $\pm $ 1.18}       &   40.49 $\pm $ 1.28     \\ 
  \quad w/o bias in loss        & \multicolumn{1}{c|}{32.33 $\pm $ 2.17}       & 44.01 $\pm $ 1.02       \\ 
    \bottomrule[1pt]
    \end{tabular}
    \end{table}
Table \ref{tab3} shows the ablation analysis of our model. The residual connection, Biaffine layer, and circle loss all contribute to the BCL's performance. Among these three components, the Biaffine layer contributes the most with about 1.73\% and 5.57\% increase in 1-shot and 5-shot, respectively. It validates the effectiveness of the span dependency for distinguishing nested entities. 
Without merging the word semantic representation to the span representation, the $F_1$ value decreases by 1.76\% and 1.29\% in 1-shot and 5-shot, respectively. Besides, the addition of the bias term $ \lambda$ in the loss function helps the model focus more on the span with the largest dissimilarity in all positive spans, leading to an increase of $F_1$ value by 1.38\% and 2.05\% in 1-shot and 5-shot, respectively. 

  \subsubsection{Influence of the Multilingual PLM}
  \begin{table}[]
    \small
    \centering
    \caption{ $F_1$ scores on the GENIA dataset with different PLMs (\%).}
    \label{tab4}
    \begin{tabular}{c|cc}
      \toprule[1pt]
      \multirow{2}{*}{PLM} & \multicolumn{2}{c}{GENIA}           \\
                             & \multicolumn{1}{c|}{1-shot} & 5-shot \\ \midrule[1pt]
      $\rm {BERT}_{base\_cased}$               & \multicolumn{1}{c|} {35.49 $\pm $ 1.09}      & 47.20 $\pm $ 1.98     \\ 
      $\rm {BERT}_{base\_multilingual}$       & \multicolumn{1}{c|}{33.71 $\pm $ 1.75}      & 46.06 $\pm $ 1.22     \\ 
      \bottomrule[1pt]
      \end{tabular}
  \end{table}
As the datasets used in our experiments are from three languages, we apply $\rm {BERT}_{base\_multilingual}$ for word embedding. To determine the influence of the multilingual PLM,  We replace $\rm {BERT}_{base\_multilingual}$ with $\rm {BERT}_{base\_cased}$. Table \ref{tab4} shows the results on the GENIA dataset. Since both the FewNERD and the GENIA are English corpus, $F_1$ scores increase a little with the English version PLM  $\rm {BERT}_{base\_cased}$ for both 1-shot and 5-shot settings. This is reasonable as training a multilingual model is more challenging. However, the difference in performance is not significant. Considering many languages lack sufficient training corpus, applying $\rm {BERT}_{base\_multilingual}$ for multiple languages is more appropriate. Note that the dataset languages used in this paper all belong to the Indo-European language family. The performance of BCL on other language families needs to be validated in the future.

\subsubsection{Results in the Flat NER Task}
\label{FlatNerTask}
We also test BCL in the flat NER task using the inter-setting and intra-setting in the FewNERD dataset \citep{ding2021few}. The results of baseline models are directly obtained from CONTaiNER \citep{das2021container}. Since the source and target domains are both English, all models utilize $\rm {BERT}_{base}$ as the PLM.
  
Table \ref{tab5} and Table \ref{tab6} show the results. We can find that BCL achieves comparable results to the state-of-the-art baseline models under 5$\sim$10 shot setting, especially BCL gets the best $F_1$ value under the inter 5-way 5$\sim$10 shot setting. When it comes to the 1$\sim$2 shot setting, the gap between our model and state-of-the-art models becomes large. The reason is that our proposed BCL framework needs more positive samples to learn the target domain entity span distribution. The baseline models are optimized in different ways when there is insufficient information in the target domain \citep{das2021container}.

  \begin{table}[]
    \scriptsize
    \centering
    \caption{ Performance on the FewNERD inter flat NER subset (\%). }
    \label{tab5}
    \begin{tabular}{c|cc|cc}
      \toprule[1pt]
    \multirow{2}{*}{Model} & \multicolumn{2}{c|}{5-way}           & \multicolumn{2}{c}{10-way}                 \\ 
                           & \multicolumn{1}{c|}{1$\sim $2 shot} & 5$\sim $10 shot & \multicolumn{1}{c|}{1$\sim $2 shot} & 5$\sim $10 shot  \\ \midrule[1pt]
    StructShot           & \multicolumn{1}{c|}{57.33}    & 57.16  & \multicolumn{1}{c|}{49.46}  & 49.39   \\ 
    ProtoNet             & \multicolumn{1}{c|}{44.44}    & 58.80  & \multicolumn{1}{c|}{39.09}  & 53.97   \\ 
    NNShot               & \multicolumn{1}{c|}{54.29}    & 50.56  & \multicolumn{1}{c|}{46.98}  & 50.00   \\ 
    CONTaiNER            & \multicolumn{1}{c|}{55.95}    & 61.83  & \multicolumn{1}{c|}{48.35}  & 57.12   \\ 
    \makecell[r]{CONTaiNER \\ \ + \ Viterbi }  & \multicolumn{1}{c|}{56.10}    & 61.90  & \multicolumn{1}{c|}{48.36}  & 57.13   \\ 
    BCL                & \multicolumn{1}{c|}{47.03}    &  63.35  & \multicolumn{1}{c|}{41.92 }      &  56.82       \\ 
    \bottomrule[1pt]
    \end{tabular}
    \end{table}

    \begin{table}[]
      \scriptsize
      \centering
      \caption{Performance on the FewNERD intra flat NER subset (\%).}
      \label{tab6}
      \begin{tabular}{c|cc|cc}
        \toprule[1pt]
      \multirow{2}{*}{Model} & \multicolumn{2}{c|}{5-way}           & \multicolumn{2}{c}{10-way}                 \\ 
                             & \multicolumn{1}{c|}{1$\sim $2 shot} & 5$\sim $10 shot & \multicolumn{1}{c|}{1$\sim $2 shot} & 5$\sim $10 shot  \\ \midrule[1pt]
      StructShot           & \multicolumn{1}{c|}{35.92}      & 38.83      & \multicolumn{1}{c|}{25.38}      & 26.39            \\ 
      ProtoNet             & \multicolumn{1}{c|}{23.45}      & 41.93      & \multicolumn{1}{c|}{19.76}      & 34.61            \\ 
      NNShot               & \multicolumn{1}{c|}{31.01}      & 35.74      & \multicolumn{1}{c|}{21.88}      & 27.67          \\ 
      CONTaiNER            & \multicolumn{1}{c|}{40.43 }  & 53.70 & \multicolumn{1}{c|}{33.84 }    & 47.49  \\ 
      \makecell[r]{CONTaiNER \\ \ + \ Viterbi }   & \multicolumn{1}{c|}{40.40 }  & 53.71 & \multicolumn{1}{c|}{33.82 }  & 47.51   \\ 
      BCL               & \multicolumn{1}{c|}{31.85}       &    50.73   & \multicolumn{1}{c|}{26.15 }       &   43.22  \\ 
      \bottomrule[1pt]
      \end{tabular}
      \end{table}

  \section{Conclusion}

This paper is the first to study the few-shot nested NER task exclusively. Facing the double challenges of nested entity structures and very few training samples, we propose a Biaffine-based contrastive learning framework to tackle this new task. We build a Biaffine span representation module to learn contextual span dependency representation that can help distinguish nested entities. Experiments on English, German, and Russian nested NER datasets validate the effectiveness of our proposed framework. This paper only scratches the surface of the few-shot nested NER task. Given that there is an overwhelming amount of unlabeled data compared to the labeled data in this task, we plan to explore ways to use the former in future work.
\bibliography{anthology}

\appendix

\end{document}